\documentclass{article}
\usepackage{spconf,amsmath,graphicx}
\usepackage{float}
\usepackage{pifont}
\usepackage{footnote}
\usepackage{enumitem}
\usepackage{bm}
\usepackage{arydshln}
\usepackage{booktabs}
\usepackage{multicol}
\usepackage{multirow}
\usepackage{color}
\usepackage{xcolor}     
\usepackage{colortbl}
\usepackage{soul}
\usepackage{bbding}
\usepackage{makecell}
\usepackage{mathtools}
\usepackage{imakeidx}
\usepackage{amssymb}
\usepackage{graphicx}
\usepackage{amsmath}
\usepackage{threeparttable}
\definecolor{citecolor}{HTML}{0071BC}
\definecolor{linkcolor}{HTML}{ED1C24}
\makeindex
\usepackage{arydshln}
\usepackage{lipsum}
\usepackage[toc]{multitoc}
\usepackage[edges]{forest}
\usepackage[normalem]{ulem}

\usepackage{bbding}
\usepackage[most]{tcolorbox}

\usepackage{algorithm}
\usepackage{algorithmic}

\usepackage{minitoc}
\usepackage[toc,page,header]{appendix}

\usepackage{enumitem}
\setlist{nosep, leftmargin=14pt}

\usepackage{mwe} 

\newcommand{\eg}{\mbox{e.g.,\ }}

\newcommand{\wrt}{\mbox{w.r.t.\ }}
\title{Large-Scale Label Quality Assessment for Medical Segmentation via a Vision-Language Judge and Synthetic Data}
\name{Yixiong Chen \qquad
Zongwei Zhou$^{\star}$\thanks{$\star$ Corresponding author: Zongwei Zhou (zzhou82@jh.edu)} \qquad
Wenxuan Li \qquad
Alan Yuille
}
\address{Johns Hopkins University, Baltimore, USA}

\begin{document}
%
\maketitle
\begin{abstract}

Large-scale medical segmentation datasets often combine manual and pseudo-labels of uneven quality, which can compromise training and evaluation. Low-quality labels may hamper performance and make the model training less robust. To address this issue, we propose SegAE (Segmentation Assessment Engine), a lightweight vision-language model (VLM) that automatically predicts label quality across 142 anatomical structures. Trained on over four million image–label pairs with quality scores, SegAE achieves a high correlation coefficient of 0.902 with ground-truth Dice similarity and evaluates a 3D mask in 0.06s. SegAE shows several practical benefits: (I) Our analysis reveals widespread low-quality labeling across public datasets; (II) SegAE improves data efficiency and training performance in active and semi-supervised learning, reducing dataset annotation cost by one-third and quality-checking time by 70\% per label. This tool provides a simple and effective solution for quality control in large-scale medical segmentation datasets.
The dataset, model weights, and codes are released.
\end{abstract}
\section{Introduction}

Recent advances in large-scale organ segmentation have shown significant potential for developing foundational models for medical segmentation. As datasets containing vast amounts of annotated scans \cite{wasserthal2023totalsegmentator,jaus2023towards,li2024abdomenatlas} have become increasingly necessary for training robust models \cite{liu2023clip}, label quality becomes a deciding factor in the scalability and effectiveness of these models.
The rapid integration of automated pseudo-labeling exacerbates this issue, resulting in inconsistent label quality across samples. This creates substantial opportunities for improving data quality and the efficiency of the training process.

The sheer scale of modern datasets, such as AbdomenAtlas \cite{li2024abdomenatlas} with its 3.7 million CT slices, makes manual quality control (QC) impractical. 
However, developing an automatic model-based QC method poses several \textbf{challenges}. Current learning‐based QC models often require extensive manual annotations \cite{bottani2022automatic} and are restricted to a few organ classes \cite{fournel2021medical}, which limits their applicability. Moreover, generic approaches based on model prediction metrics like entropy \cite{culotta2005reducing} or prediction inconsistency \cite{gal2017deep} do not reliably capture the true quality of segmentation labels and suffer from suboptimal accuracy and efficiency. 
These challenges motivate our work to address three key issues: the desire to reduce reliance on manual annotations, the need for a scalable QC model across a diverse range of organs, and the requirement for accurate and efficient quality assessment.

To address the above-mentioned issues, we devise a learning-based framework that trains QC models on a synthetic quality dataset. This framework, \textbf{SegAE}, short for \textbf{Seg}mentation \textbf{A}ssessment \textbf{E}ngine, only relies on the label-quality pairs generated in the training process of a segmentation model. It uses BiomedCLIP encoders \cite{zhang2023biomedclip} to learn from the Dice Similarity Coefficient (DSC) between pseudo-labels and ground truth as a predictive signal for label quality. 
What sets SegAE apart from existing QC methods \cite{fournel2021medical,zhang2021quality,bottani2022automatic,zhou2023volumetric} is it introduces a vision language model (VLM) as a judge to encode not only the image-label pairs, but also the label semantic via text. This multi-modal input allows the model to be aware of the anatomical structure it is evaluating, scaling up the QC classes in complex human bodies.
SegAE is accurate and efficient, it reaches a high positive correlation coefficient ($r=0.902$) with ground truth DSC and can assess a 3D label in 0.06 s on a Nvidia A6000 GPU.

In summary, our work presents three major \textbf{contributions} to the field of medical image segmentation. First, we introduce SegAE, a novel vision-language QC model designed to efficiently assess 3D segmentation label quality through DSC prediction (Sect. \ref{sect:method}). Second, we validate that SegAE can predict highly accurate quality scores for both in-distribution and out-of-distribution data. Our external evaluation also reveals that up to 10\% of the masks from the existing large-scale datasets are poor in quality. (Sect. \ref{sect:eval} and Sect. \ref{sect:external}). Third, we demonstrate the efficacy of SegAE as a sample selector for training segmentation models. It outperforms traditional QC methods for both active learning and semi-supervised training (Sect. \ref{sect:efficient_training}).

\section{Segmentation Assessment Engine}
\label{sect:method}

\begin{figure}[t]
\vspace{0cm}                          
\centering\centerline{\includegraphics[width=1.0\linewidth]{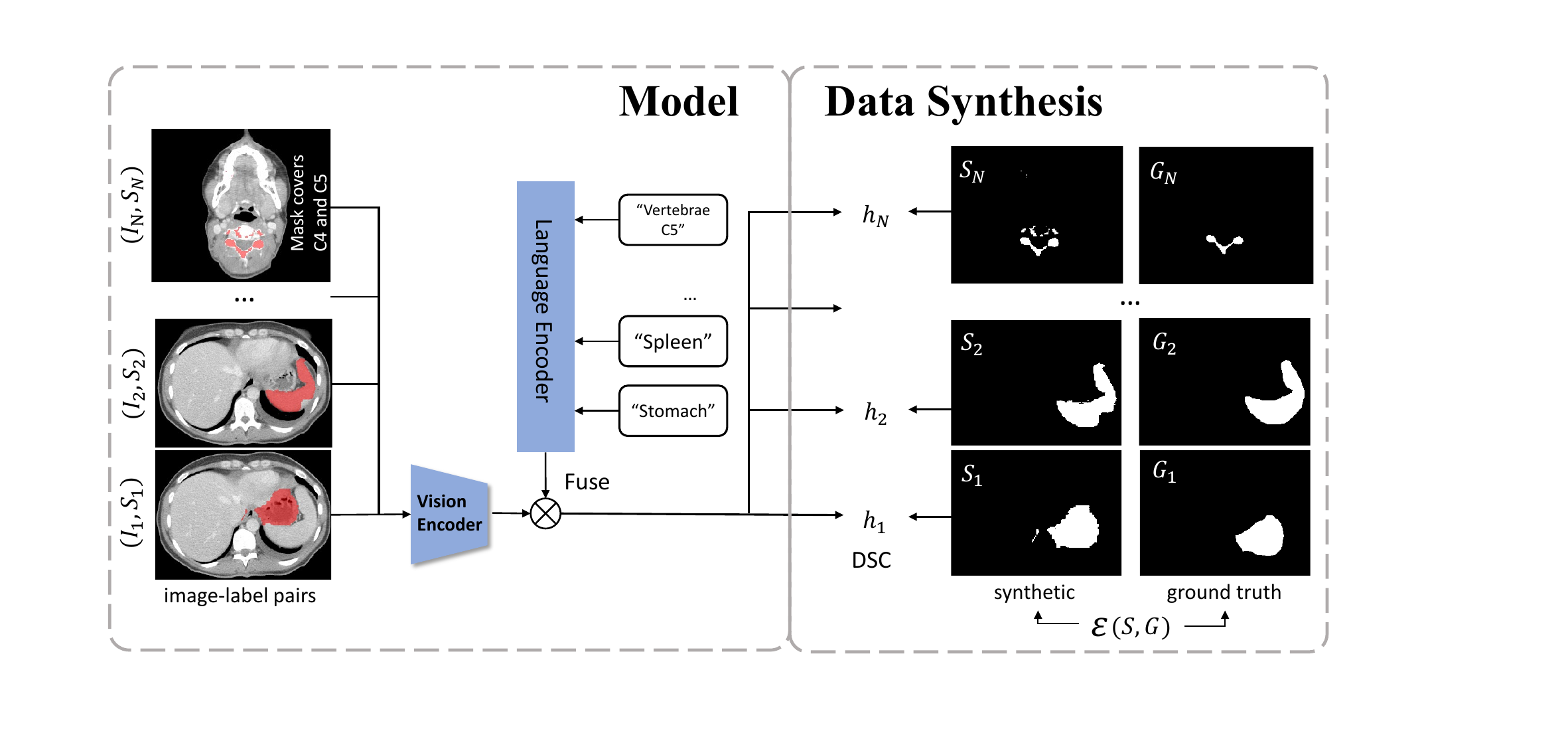}}
\caption{Overview of SegAE: SegAE employs a vision encoder to encode an image with its visualized segmentation mask, and a language encoder to encode the text description of a target class. It predicts DSC relative to ground truths. Training is performed on synthetic masks generated from segmentation checkpoints.}
\label{fig:SegAE}
\end{figure}

\subsection{Formulation of the Framework}
The SegAE framework, depicted in Fig. \ref{fig:SegAE}, evaluates data batches of $N$ samples, denoted as  $\mathcal{D}  =\{I_i, S_i, T_i, G_i\}_{i=1}^N$, where $I_i$, $S_i$, $T_i$, and $G_i$ represent an image, a segmented mask (pseudo-label), a text description, and a ground-truth label, respectively. The assessment model $\mathcal{Q}$, parameterized by $\theta$, predicts the DSC $\hat{h}_i = \mathcal{Q}(I_i, S_i, T_i; \theta)$, compared against the actual DSC $h_i = \mathcal{E}(S_i, G_i)$. To align the predicted $\hat{h}$ with the actual $h$, we introduce a compositional loss function:
\begin{equation}
    \mathcal{L}=\frac{1}{N}\sum_{i=1}^N\mathcal{L}_{MSE}(\hat{h}_i, h_i)+\lambda \mathcal{L}_{Rank}(\hat{\textbf{h}}, \textbf{h}),
\label{eq1}
\end{equation}
with $\lambda$ as a scaling constant. The mean square error (MSE) loss $\mathcal{L}_{MSE}(\hat{h}_i, h_i)$ aims to match the model's output with the ground truth DSC value. The novel optimal-pair ranking loss $\mathcal{L}_{Rank}(\hat{\textbf{h}}, \textbf{h})$ enhances the model's ranking capability by optimizing the order of predicted and actual DSC within the batch.

\subsection{SegAE Architecture}
Recent QC methods \cite{fournel2021medical,bottani2022automatic} are typically designed for specific organs. Extending them to large-scale segmentation introduces ambiguity in organ–label correspondence, especially for overlapping structures (\eg adjacent vertebrae). As shown in the upper left of Fig. \ref{fig:SegAE}, a mask may cover adjacent vertebrae, creating ambiguity in organ-label matching and compromising model accuracy. This ambiguity compromises the QC model's accuracy. To address this, SegAE uses BiomedCLIP's \cite{zhang2023biomedclip} language encoder to encode the text semantics of the target class to discern the target class amidst multiple classes within an image. The text embeddings also encourage the model to learn shared representations for similar classes (\eg left and right kidneys) due to their semantic similarity. In this way, the shared assessment criteria can promote each other's performance.

SegAE fuses image features and class text embeddings with a class-conditional attention before the decision layers. For a given class $k$, its text embedding $\varphi_k$ is concatenated with the image feature $f_1 = E_v(x)$ extracted by a vision encoder. A multi-layer perceptron (MLP) then computes attention weights $[\omega_1, \omega_2] = MLP([f_1, \varphi_k])$, which are used by the regression head consisting of two fully connected (FC) layers $g_1(\cdot)$ and $g_2(\cdot)$ to produce the final prediction:
\begin{equation}
\hat{h} = g_2(\omega_2 \otimes g_1(\omega_1 \otimes E_v(x))),
\label{eq2}
\end{equation}
where $\otimes$ denotes element-wise multiplication. The two FC layers enable progressive feature compression, allowing the class-conditional attention to exert a stronger influence on predictions.

\subsection{Optimal Pair Ranking Loss}
Training SegAE with only MSE loss is suboptimal, as it biases DSC predictions toward the most frequent score range. Given that SegAE is designed to assess label quality and rank them, we introduce a scale-insensitive ranking loss as an auxiliary objective.

Inspired by \cite{yoo2019learning}, we adopt a pairwise comparison approach. However, in multi-class segmentation, comparing unrelated classes (\eg spleen vs. skull) is ineffective. To address this, we construct $N/2$ pairs within a batch of size $N$, ensuring each pair consists of identical or similar classes by maximizing embedding similarity. The pairing process follows three steps: (1) computing the cosine similarity matrix $H \in \mathbb{R}^{N\times N}$, (2) converting $H$ into a cost matrix $H' = -H + \infty \cdot I_{N\times N}$ to prevent self-pairing, and (3) using the Jonker-Volgenant algorithm \cite{crouse2016implementing} to determine optimal pairs.

For an optimal pair $\{x^p=\{x_i, x_j\}\}$, the ranking loss is defined as:
\begin{equation}
    \mathcal{L}_{Rank}(\hat{h}_i,\hat{h}_j,h_i,h_j) = \max(0, (\hat{h}_i-\hat{h}_j)\cdot (h_j-h_i) + \xi),
\label{eq3}
\end{equation}
where $\xi$ is a predefined positive margin. This loss penalizes incorrect directional relationships between predicted and actual DSC differences, encouraging better ranking alignment.

\subsection{Label Quality Data Synthesis}

\textbf{Stage 1.} We fine-tuned STUNet \cite{huang2023stu} on the DAP Atlas \cite{jaus2023towards}, a large-scale CT segmentation dataset with 142 organs. The dataset includes diverse CT scans from multiple hospitals, acquired using standardized PET/CT scanners (Siemens Biograph mCT, mCT Flow, Biograph 64, GE Discovery 690). Pseudo-labels were generated from different model checkpoints in training process on 10 3D CT scans, covering a whole range of DSC values. To balance the label distribution, we generated more samples at earlier epochs: ${10, 20, 30, 40, 50, 100, 200, 300, 400, 500}$. Each 3D CT produces multiple 2D slices paired with intermediate pseudo-labels from different checkpoints, yielding over one million slice–mask pairs. This construction pipeline avoids manual annotations like previous works \cite{bottani2022automatic}.

\textbf{Stage 2.} After training on stage 1 data, SegAE resampled 50 3D CT scans with high-quality labels from the DAP Atlas, followed by human screening. This stage ensures balanced sex distribution and 4 million slice-mask pairs with high precision, leading to more accurate DSC predictions and better model performance. We do 8:2 train-test CT-wise split. The label quality is verified in our ablation study (Sect. \ref{sect:ablation}) and dataset quality benchmark (Sect. \ref{sect:data_eval}).

\begin{table}[t]
\small
\centering
\caption{Ablations on class conditioning, ranking loss, and resampling. Results on the resampled test split.}
\renewcommand{\arraystretch}{1.1}
\resizebox{\linewidth}{!}{%
\begin{tabular}{p{5.3cm}cccccc}
\toprule
Condition & Opt.~Pair & Resample & LCC & SROCC & MAP@5 & MAP@10 \\
\midrule
Class one-hot & - & - & 0.817 & 0.795 & 0.427 & 0.501 \\
text: ``A CT of a [CLS].'' & - & - & 0.840 & 0.805 & 0.438 & 0.510 \\
text: ``There is [CLS] in this CT.'' & - & - & 0.847 & 0.801 & 0.444 & 0.525 \\
text: ``A photo of a [CLS].'' & - & - & 0.852 & 0.808 & 0.445 & 0.533 \\
text: ``[CLS]'' (used) & - & - & 0.853 & 0.808 & 0.449 & 0.531 \\
\midrule
- & - & - & 0.797 & 0.775 & 0.417 & 0.481 \\
\checkmark & - & - & 0.853 & 0.808 & 0.449 & 0.531 \\
- & \checkmark & - & 0.812 & 0.799 & 0.438 & 0.535 \\
\checkmark & \checkmark & - & 0.875 & 0.820 & 0.482 & 0.562 \\
\checkmark & \checkmark & \checkmark & \textbf{0.902} & \textbf{0.856} & \textbf{0.500} & \textbf{0.565} \\
\bottomrule
\end{tabular}}
\label{tab:ablation}
\end{table}

\begin{figure}[t]
\vspace{0cm}                          
\centering\centerline{\includegraphics[width=1.0\linewidth]{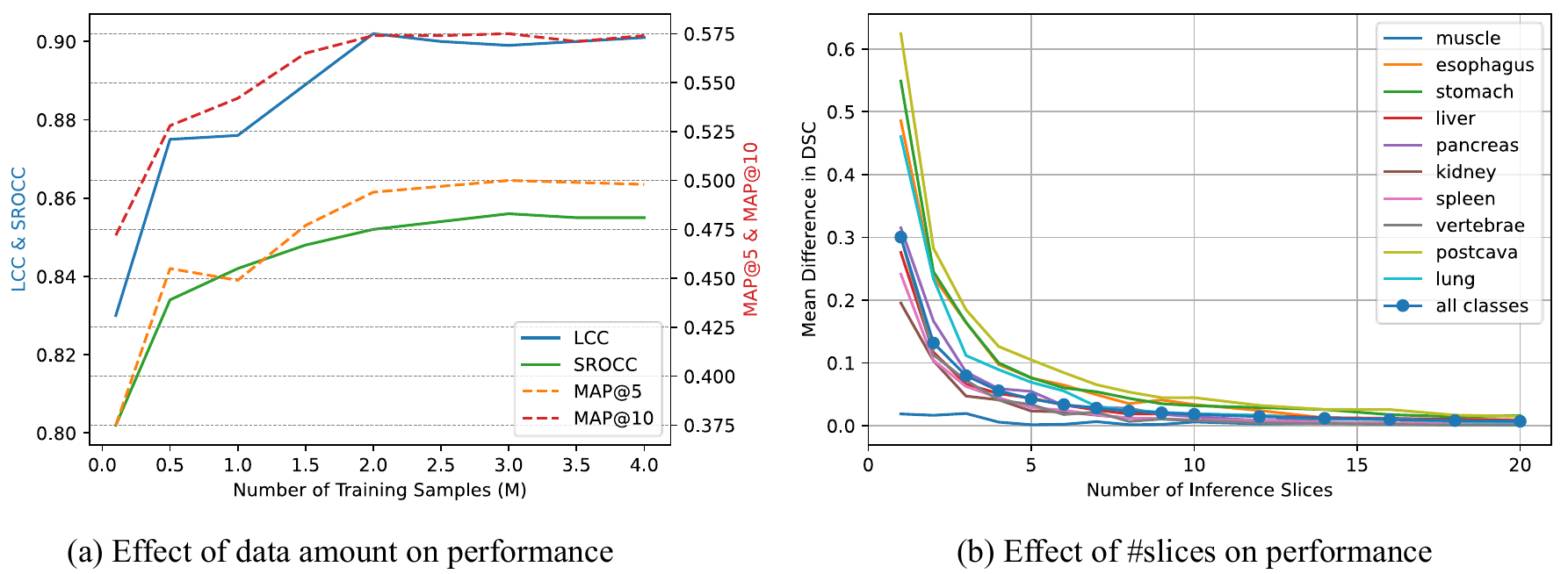}}
\caption{Illustration of (a) the curves of model LCC/SROCC and data amount and (b) the curves of performance deviation \wrt \#slice compared to prediction with all slices.}
\label{fig:effect}
\end{figure}

\section{Experiments \& Results}

\subsection{Experimental Setting}
We used BiomedCLIP's \cite{zhang2023biomedclip} pretrained vision and language encoder to implement SegAE. Text embeddings (512D) were precomputed for efficiency. CT HU values were clipped to $[-200,200]$, images were cropped and resized to $256\times 256$, then concatenated with pseudo-label masks. SegAE model was trained for 30 epochs using AdamW optimizer ($\text{lr}=10^{-3}$, batch size 128, $\lambda=1$). Experiments ran on PyTorch 2.9.0 with an Intel Xeon Gold 5218R CPU and 8 Nvidia RTX A6000 GPUs.

\subsection{Evaluation of SegAE}
\label{sect:eval}

\subsubsection{Ablation Study.}
\label{sect:ablation}
Table~\ref{tab:ablation} shows that class conditioning improves correlation and ranking; text embeddings outperform one-hot, with the simplest prompt ``[CLS]'' giving the best results. 
The optimal-pair ranking term improves ranking (MAP@10 from $0.531$ to $0.562$). 
Resampling further boosts all metrics; the full model Correlation Coefficient (LCC) $0.902$ and Spearman Rank Order Correlation Coefficient (SROCC) $0.856$.

\subsubsection{Evaluation of Model Configurations.} 
Performance plateaus at $2$–$3$M training pairs (Fig.~\ref{fig:effect}a). 
For 3D masks, averaging predictions over $10$ uniformly sampled slices provides accurate estimates with small deviation from full-slice scoring (Fig.~\ref{fig:effect}b).

\begin{figure}[t]
\vspace{0cm}                          
\centering\centerline{\includegraphics[width=1.0\linewidth]{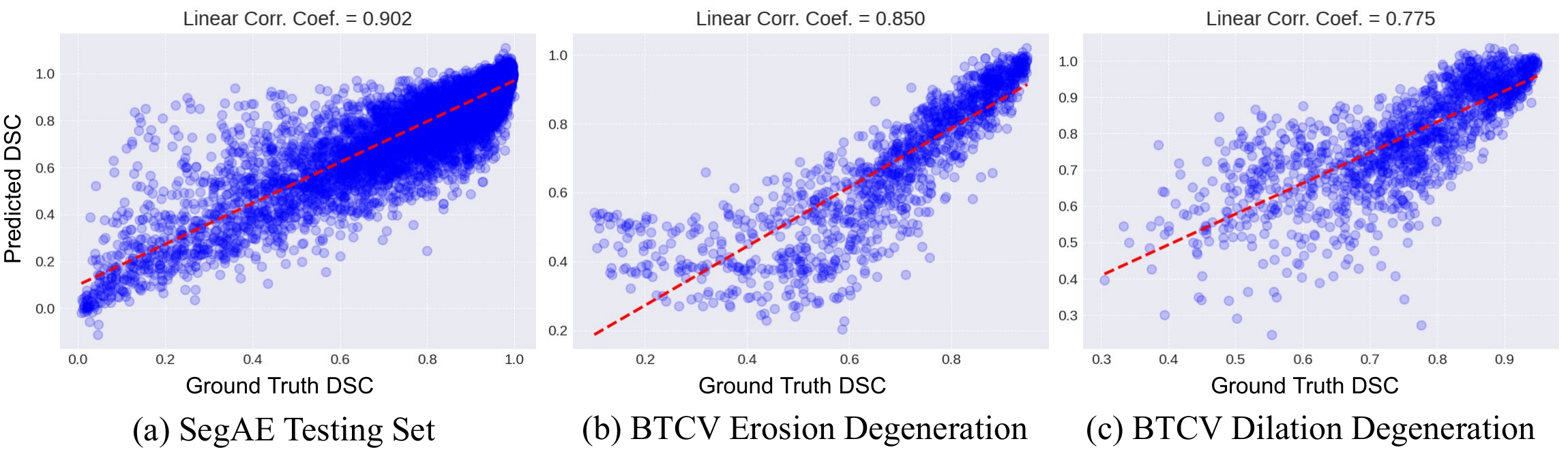}}
\caption{The scatter plots between predicted DSC and ground truth DSC of (a) SegAE testing set and (b, c) external evaluation on BTCV with two degenerations.}
\label{fig:scatter}
\end{figure}

\begin{figure}[t]
\vspace{0cm}                          
\centering\centerline{\includegraphics[width=1.0\linewidth]{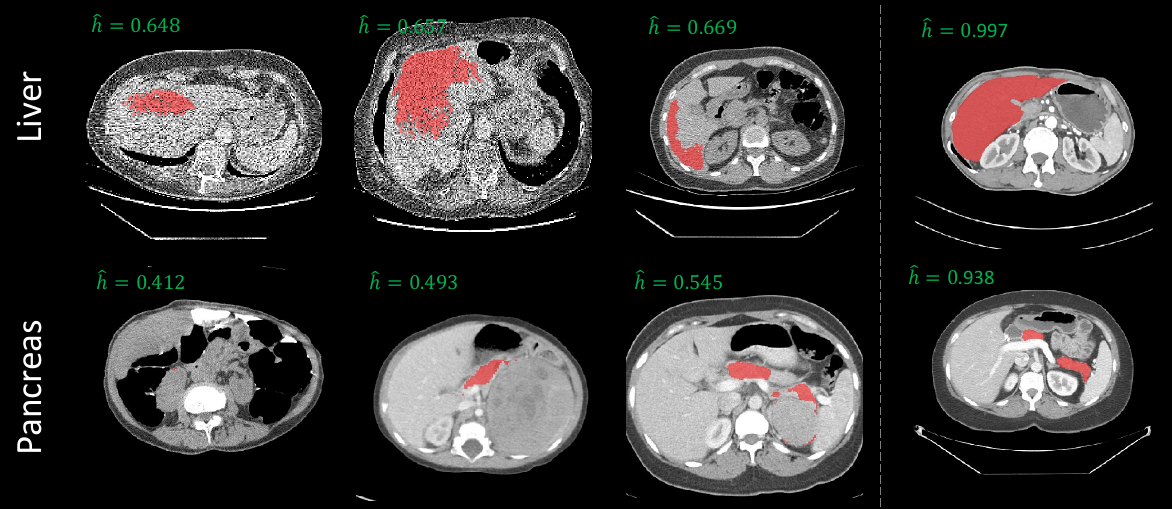}}
\caption{Illustration of samples with bad (left) and good (right) labels in AbdomenAtlas dataset.}
\label{fig:AAtlas}
\end{figure}

\subsubsection{External Evaluation.}
\label{sect:external}
On the held-out test set SegAE closely matches ground truth (Fig.~\ref{fig:scatter}a). 
On BTCV~\cite{landman2015miccai}, synthetic degradations (erosion/dilation) are ranked reliably with $LCC=0.850/0.775$ (Fig.~\ref{fig:scatter}b–c), indicating robustness to distribution shift.

\subsection{Dataset Benchmarking with SegAE}
\label{sect:data_eval}

\begin{table*}[t]
  \centering
  \caption{Label Quality (Predicted DSC) benchmark of different multi-organ segmentation datasets evaluated by SegAE.}
  \label{tab:eval}
  \resizebox{\textwidth}{!}{%
  \begin{tabular}{l|c|c|cccccccc|cc}
    \toprule
    ~~~~~~Dataset & \# Samples & Annotators & \multicolumn{8}{c|}{Organ-wise Average Quality (Predicted DSC)} & \multicolumn{2}{c}{Overall Quality}  \\
    \cline{4-13}
      & & & aorta & gallbladder & kidney & liver & pancreas & postcava & spleen & stomach & Mean DSC~ & DSC$<0.8$  \\
    \hline
    \hline
    BTCV & 47 & Human &  0.943 & 0.875 & 0.961 & \textbf{0.981} & 0.928 & 0.921 & 0.962 & \textbf{0.969} & 0.944  & 1.7\% \\
    AbdomenCT-12organ & 50 & Human & 0.965 & \textbf{0.932} & 0.963 & 0.965 & \textbf{0.961} & \textbf{0.960} & 0.964 & 0.961 & 0.959 & \textbf{0.0\%} \\
    AMOS22 & 200 & Human\&AI & 0.954 & 0.874 & 0.956 & 0.971 & 0.928 & 0.927 & 0.953 & 0.942 & 0.941 & 2.1\% \\
    DAP Atlas & 533 & AI & 0.945 & 0.856 & 0.946 & 0.961 & 0.851 & 0.919 & 0.959 & 0.945 & 0.925 & 2.4\% \\
    TotalSegmentator & 1204 & Human\&AI & 0.927 & 0.794 & 0.914 & 0.941 & 0.853 & 0.880 & 0.933 & 0.915 & 0.901 & 8.0\% \\
    AbdomenAtlas & 5195 & Human\&AI & 0.884 & 0.764 & 0.923 & 0.948 & 0.885 & 0.842 & 0.934 & 0.873 & 0.886 & 12.6\% \\
    \hline
    DAP Atlas resampled & 50 & Human\&AI & \textbf{0.968} & 0.911 & \textbf{0.972} & 0.980 & 0.944 & \textbf{0.960} & \textbf{0.983} & 0.959 & \textbf{0.964} & \textbf{0.0\%} \\
    \bottomrule
  \end{tabular}%
  }
\end{table*}

\begin{table*}[t]
\centering
\scriptsize
\caption{Performance and efficiency comparison on semi-supervised learning. We conduct experiments on TotalSegmentator for organs and AbdomenAtlas-Tumors for unseen tumors.}
\label{tab:semi}
\setlength{\tabcolsep}{3.0pt}
\renewcommand{\arraystretch}{1.05}
\resizebox{\textwidth}{!}{%
\begin{tabular}{l cc cc ccc ccc}
\toprule
& \multicolumn{2}{c}{\# Scans Used}
& \multicolumn{2}{c}{TotalSegmentator}
& \multicolumn{3}{c}{AbdomenAtlas-Tumors: Detection Acc.}
& \multicolumn{3}{c}{Quality Assessment Cost} \\
\cmidrule(lr){2-3}\cmidrule(lr){4-5}\cmidrule(lr){6-8}\cmidrule(lr){9-11}
Method & Labeled & Unlabeled
& DSC [\%] & NSD [\%]
& Liver [\%] & Kidney [\%] & Pancreas [\%]
& Time [s/label] & RAM [GB] & Disk [GB] \\
\midrule
SwinUNETR & 20 & 0
& 82.6{\tiny$\pm$0.3} & 45.8{\tiny$\pm$0.3}
& 67.5{\tiny$\pm$1.4} & 73.9{\tiny$\pm$0.9} & 64.6{\tiny$\pm$1.0}
& -- & -- & -- \\
Random & 20 & 100
& 83.2{\tiny$\pm$0.3} & 46.1{\tiny$\pm$0.3}
& 72.7{\tiny$\pm$1.9} & 79.4{\tiny$\pm$1.2} & 70.1{\tiny$\pm$2.2}
& -- & -- & -- \\
MC dropout~\cite{gal2016dropout} & 20 & 100
& 83.3{\tiny$\pm$0.6} & 46.4{\tiny$\pm$0.5}
& 74.2{\tiny$\pm$1.1} & 80.4{\tiny$\pm$1.4} & 71.1{\tiny$\pm$1.4}
& 0.42 & 92.4 & 1032 \\
Entropy~\cite{joshi2009multi} & 20 & 100
& 83.2{\tiny$\pm$0.4} & 46.2{\tiny$\pm$0.4}
& 73.9{\tiny$\pm$0.8} & 81.9{\tiny$\pm$0.7} & 70.5{\tiny$\pm$1.1}
& 0.20 & 33.1 & 344 \\
SegAE (ours) & 20 & 100
& \textbf{84.8}{\tiny$\pm$0.4} & \textbf{48.1}{\tiny$\pm$0.4}
& \textbf{77.6}{\tiny$\pm$0.8} & \textbf{83.6}{\tiny$\pm$0.7} & \textbf{73.1}{\tiny$\pm$1.3}
& 0.06 & 1.5 & 0.05 \\
\bottomrule
\end{tabular}}
\end{table*}

SegAE highlights typical issues such as reduced, missing, and expanded masks (Fig.~\ref{fig:AAtlas}) in AbdomenAtlas~\cite{li2024abdomenatlas}.
We conducted a quantitative benchmarking of 6 segmentation datasets focusing on major abdominal organs using SegAE, as detailed in Tab. \ref{tab:eval}. We observed that smaller human-annotated datasets, exhibit superior organ-wise label quality. Conversely, in larger datasets like TotalSegmentator and AbdomenAtlas, there is a noticeable decline in the overall DSC estimation and more low quality labels with $DSC<0.8$. SegAE reveals $8-13\%$ low-quality labels (DSC < 0.8) in large mixed-source datasets, motivating further re-annotation. After SegAE's filtering and human screening, DAP Atlas presents a large improvement on label quality $Mean =0.964,~0\% ~DSC<0.8$.

\begin{figure}[t]
\vspace{0cm}                          
\centering\centerline{\includegraphics[width=1.0\linewidth]{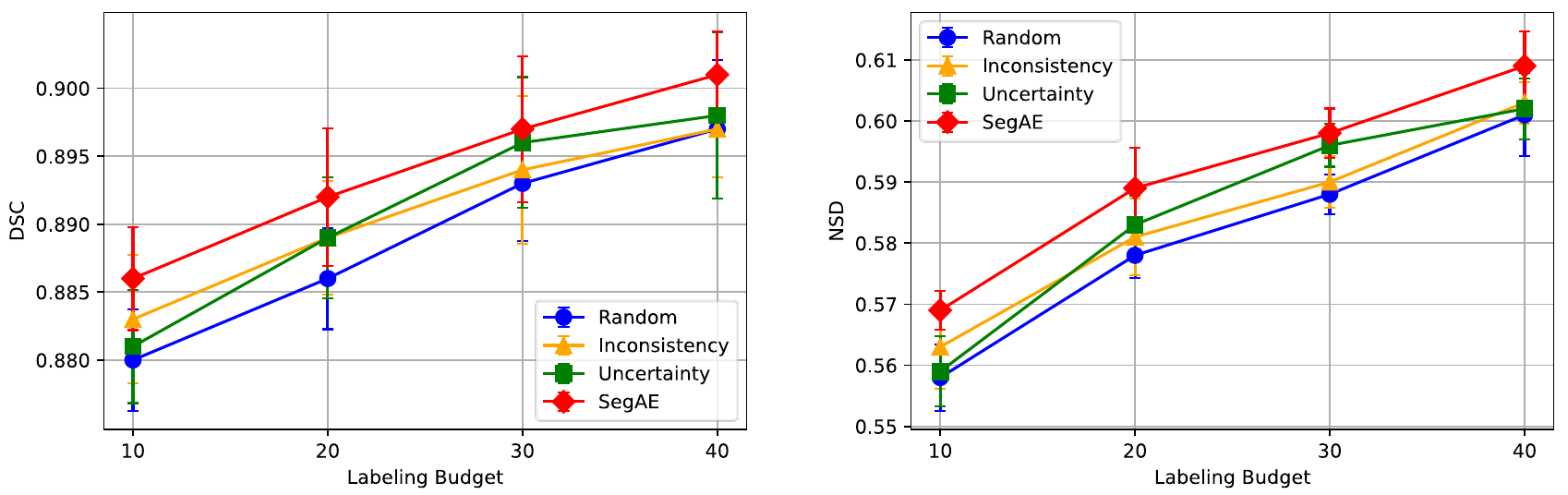}}
\caption{Comparison of QC methods on active learning.}
\label{fig:active}
\end{figure}

\subsection{Data-Efficient Training}
\label{sect:efficient_training}
We assess whether SegAE improves sample selection on TotalSegmentator~\cite{wasserthal2023totalsegmentator} and AbdomenAtlas-Tumors \cite{li2024abdomenatlas} using a pretrained SwinUNETR~\cite{tang2022self}. 
We fine-tune with $20$ labeled full-body scans and $100$ pseudo-labeled candidates. 
Baselines are random, Monte-Carlo dropout~\cite{gal2016dropout}, and entropy~\cite{joshi2009multi}. 
Metrics are DSC and NSD (five trials).

\textbf{Active learning.}
Selecting the $n$ lowest-quality cases in TotalSegmentator for manual correction, SegAE yields consistent gains (Fig.~\ref{fig:active}). 
With a labeling budget of $20$ cases, improvements over random are $+0.007$ DSC ($p{=}0.0065$) and $+0.011$ NSD ($p{=}0.0127$), reaching the performance of annotating 30 random cases while annotating only 20.

\textbf{Semi-supervised learning with unseen tumor classes.}
Selecting the top-quality $20$ pseudo-labels and mixing them with $20$ labeled scans improves performance beyond uncertainty-based selection (Table~\ref{tab:semi}). Relative gains over entropy are $+1.6$ points DSC and $+1.9$ points NSD on TotalSegmentator.
Moreover, SegAE generalizes to unseen tumor categories using only their textual names as prompts on AbdomenAtlas-Tumors, achieving reliable ranking without any re-training.

\textbf{Efficiency.}
SegAE evaluates one 3D mask in $\approx0.06$\,s on an RTX A6000 and needs far less RAM and disk than entropy and MC-dropout (Table~\ref{tab:semi}) because it operates on 2D image–mask pairs and avoids heavy computation and I/O on 3D volumetric probability maps.

\section{Conclusion}

Our primary contribution is the development of SegAE, a vision-language QC model that was trained on a large-scale synthetic quality dataset. This model enables thorough assessment of label quality, and acts as a pivotal tool for dataset benchmarking and sample selection, both for guiding active and semi-supervised model training. While our empirical results show the practical value of the model on CT images, an important future work is to extend SegAE to more modalities to be a generic label evaluator for medical image segmentation. We envision that SegAE will be a simple and efficient tool in QC for expansive datasets.

\section{Statement on Ethical Compliance}
There is no ethical approval required for this work.

\section{Acknowledgments}
This work was supported by the Lustgarten Foundation for Pancreatic Cancer Research and the National Institutes of Health (NIH) under Award Number R01EB037669. Codes and models are available at https://github.com/Schuture/SegAE.

\bibliographystyle{IEEEbib}
\bibliography{SegAE}

\end{document}